\documentclass[conference,9pt]{IEEEtran}
\IEEEoverridecommandlockouts
\usepackage{cite}
\usepackage{amsmath,amssymb,amsfonts}
\usepackage{algorithmic}
\usepackage{graphicx}
\usepackage{textcomp}
\usepackage{xcolor}
\usepackage{threeparttable}
\usepackage{tabularx}
\usepackage{booktabs}
\usepackage{caption}
\usepackage{colortbl}
\usepackage{subcaption}
\usepackage{pifont}
\usepackage[capitalise]{cleveref}
\usepackage{multirow}
\usepackage[normalem]{ulem}

\crefname{figure}{Fig.}{Fig.}
\Crefname{figure}{Fig.}{Fig.}
\crefname{algorithm}{Algorithm}{Algorithm}
\Crefname{algorithm}{Algorithm}{Algorithm}
\crefdefaultlabelformat{#2(#1)#3}

\usepackage[ruled,vlined,linesnumbered]{algorithm2e}
\def\BibTeX{{\rm B\kern-.05em{\sc i\kern-.025em b}\kern-.08em
    T\kern-.1667em\lower.7ex\hbox{E}\kern-.125emX}}
\begin{document}

\title{SparseTrain: Exploiting Dataflow Sparsity for Efficient Convolutional Neural Networks Training
}
\author{Pengcheng Dai\IEEEauthorrefmark{1},
Jianlei Yang\IEEEauthorrefmark{2},
Xucheng Ye\IEEEauthorrefmark{2},
Xingzhou Cheng\IEEEauthorrefmark{2},
Junyu Luo\IEEEauthorrefmark{2},
Linghao Song\IEEEauthorrefmark{3},\\
Yiran Chen\IEEEauthorrefmark{3} and Weisheng Zhao\IEEEauthorrefmark{1}\\
\IEEEauthorblockA{\IEEEauthorrefmark{1}School of Microelectronics, BDBC, Beihang University, Beijing, China}
\IEEEauthorblockA{\IEEEauthorrefmark{2}School of Computer Science and Engineering, BDBC, Beihang University, Beijing, China}
\IEEEauthorblockA{\IEEEauthorrefmark{3}Department of Electrical and Computer Engineering, Duke University, Duram, NC, USA}
Email: jianlei@buaa.edu.cn~~weisheng.zhao@buaa.edu.cn
\thanks{This work is supported in part by the National Natural Science Foundation of China (61602022), State Key Laboratory of Software Development Environment (SKLSDE-2018ZX-07), CCF-Tencent IAGR20180101 and the 111 Talent Program B16001.}
}

\maketitle

\begin{abstract}

Training Convolutional Neural Networks (CNNs) usually requires a large number of computational resources.
In this paper, \textit{SparseTrain} is proposed to accelerate CNN training by fully exploiting the sparsity. It mainly involves three levels of innovations: activation gradients pruning algorithm, sparse training dataflow, and accelerator architecture. By applying a stochastic pruning algorithm on each layer, the sparsity of back-propagation gradients can be increased dramatically without degrading training accuracy and convergence rate. Moreover, to utilize both \textit{natural sparsity} (resulted from ReLU or Pooling layers) and \textit{artificial sparsity} (brought by pruning algorithm), a sparse-aware architecture is proposed for training acceleration. This architecture supports forward and back-propagation of CNN by adopting 1-Dimensional convolution dataflow. We have built 
a cycle-accurate architecture simulator to evaluate the performance and efficiency based on the synthesized design with $14nm$ FinFET technologies.
Evaluation results on AlexNet/ResNet show that \textit{SparseTrain} could achieve about $2.7 \times$ speedup and $2.2 \times$ energy efficiency improvement on average compared with the original training process.


\end{abstract}


\section{Introduction} \label{section-1}

Recent years, Convolutional Neural Networks (CNNs) have been widely used in computer vision tasks such as image classification \cite{krizhevsky2012imagenet}, object detection \cite{ren2015faster}, face recognition \cite{sun2015deepid3}, and object tracking \cite{bertinetto2016fully}.
Modern CNN models could achieve state-of-the-art accuracy by extracting rich features and significantly outperform traditional methods \cite{lecun2015deep}.
However, running large scale CNN models often requires large memory space and consumes huge computational resources.

Many CNN accelerator architectures have been invented to improve the CNN inference throughput and energy efficiency in both industrial \cite{jouppi2017datacenter, jouppi2018motivation} and academic \cite{chen2016eyeriss, parashar2017scnn, imani2019floatpim, he2018joint, li2018network, ren2019admm, wen2017coordinating, wen2017learning, chen2019slide} communities.
Among them, exploiting network sparsity has been observed as one of the most efficient approaches.
For example, by applying weight pruning, the model size of AlexNet and VGG-16 could be reduced by $35\times$ and $49\times$, respectively \cite{han2015deep}, which provides a huge potential for accelerating CNN inference.

\begin{figure}[ht]
  \centering
  \begin{subfigure}[t]{6.85pc}
    \centering
    \includegraphics{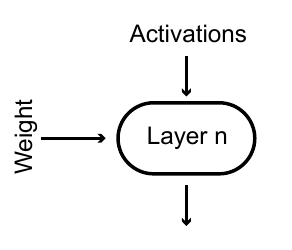}
    \caption{Inference without sparsity.} \label{fig-1-1:no-sparse}
  \end{subfigure}
  \begin{subfigure}[t]{13.85pc}
    \centering
    \includegraphics{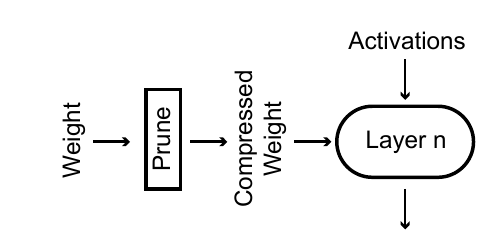}
    \caption{Inference with weight sparsity.} \label{fig-1-2:weight-sparse}
  \end{subfigure}
  \par
  \begin{subfigure}[t]{6.85pc}
    \centering
    \includegraphics{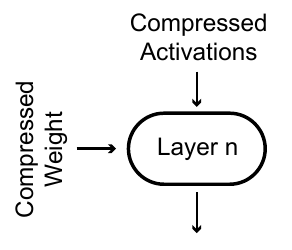}
    \caption{Inference with activation sparsity.} \label{fig-1-3:activation-sparse}
  \end{subfigure}
  \begin{subfigure}[t]{13.85pc}
    \centering
    \includegraphics{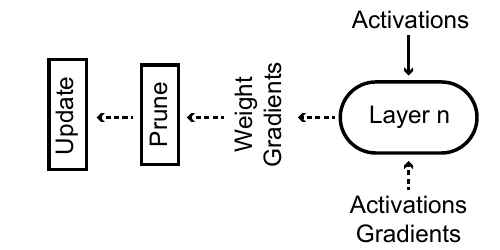}
    \caption{Training with weight gradient sparsity.} \label{fig-1-4:wg-sparse}
  \end{subfigure}
  \par
  \begin{subfigure}[t]{16pc}
    \centering
    \includegraphics{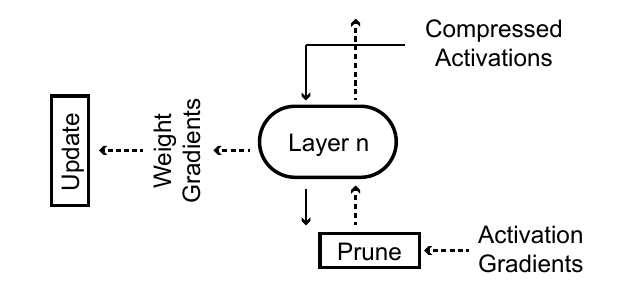}
    \caption{Ours: training with activation gradient sparsity.} \label{fig-1-5:ours}
  \end{subfigure}
  \caption{CNN sparsity exploration.} \label{fig-format}
  \vspace{-1.2pc}
\end{figure}

In fact, by using weight sparsity, a sparse-aware computing architecture --- EIE \cite{han2016eie} achieved $13\times$ speedups and $3400\times$ energy efficiency when compared to GPU implementation of the same DNN without compression.
\Cref{fig-1-1:no-sparse} and \cref{fig-1-2:weight-sparse} show the original CNN inference process and the one with weight pruning and sparsity utilization, respectively.
One step further, SCNN \cite{parashar2017scnn} achieved significant speedup in CNNs inference by utilizing both weight sparsity and \textit{natural sparsity} of activations. \Cref{fig-1-3:activation-sparse} shows the basic idea of this procedure.

Most works focused on the inference of CNN. However, the sparsity of the training process was less studied.
Compared with inference, CNN training demands much more computational resources. Usually CNN training
introduces about $3 \times$ computational cost and consumes $10\times$ to $100 \times$ memory space compared with the inference.

To utilize gradient sparsity in CNNs training, \cite{wen2017terngrad}\cite{he2019simultaneously} proposed a method that uses only three numerical levels \{-1,0,1\} for weight gradients to reduce communication in distributed training, which can be considered as a mixture of pruning and quantization. This scheme is shown in \cref{fig-1-4:wg-sparse}. However, it can hardly reduce the computation since the main training process (forward and back-propagation) is still calculated in a dense data format. There are also many training accelerators try to utilizing sparsity, which can be found in \cite{comparison}. But neither of them exploit gradient sparsity.

Since activation gradients is an operand for both backward process and weight gradients generation process, the sparsity of activation gradients provides a significant reduction on the computation of training. To overcome the limitation of previous works and fully exploit the sparsity of CNN training, we proposed a layer-wise gradients pruning method that can improve the training performance remarkably by generating \textit{artificial sparsity} on activation gradients data. Our scheme is shown in \cref{fig-1-5:ours}. The key contributions of this paper are:

\begin{itemize}
    \item We present a layer-wise gradients pruning algorithm that can generate high artificial sparsity on activation gradients data with negligible overhead. Our experiments show that this algorithm will hardly degrade the accuracy and convergence rate of CNNs.

    \item We propose a 1-Dimensional Convolution based dataflow. Combined with our pruning algorithm, it can fully utilize all kinds of sparsity in training.

    \item To support our sparse dataflow, an accelerator architecture is designed for CNNs training that can benefit from both activation and gradient sparsity.
\end{itemize}

The proposed \textit{SparseTrain} scheme is evaluated by AlexNet \cite{krizhevsky2012imagenet} and ResNet \cite{he2016deep} on CIFAR/ ImageNet.
Evaluation results on AlexNet with natural sparsity show \textit{SparseTrain} achieves $2.7 \times$ speedup and $2.2 \times$ energy efficiency on average, compared with the dense baseline.

The remaining of this paper is organized as follows.
Section \ref{section-2} introduces CNN training basics.
Section \ref{section-3} provides a gradients pruning algorithm, as well as the threshold determination and prediction method.
The details of our sparse training dataflow are presented in Section \ref{section-4}.
Section \ref{section-5} demonstrates the sparse-aware architecture designed for the proposed training dataflow.
Evaluation results are discussed in Section \ref{section-6}.
Section \ref{section-7} concludes this paper.

\section{Preliminary} \label{section-2}


\begin{figure}[t]
  \centering
  \includegraphics[width=1\columnwidth]{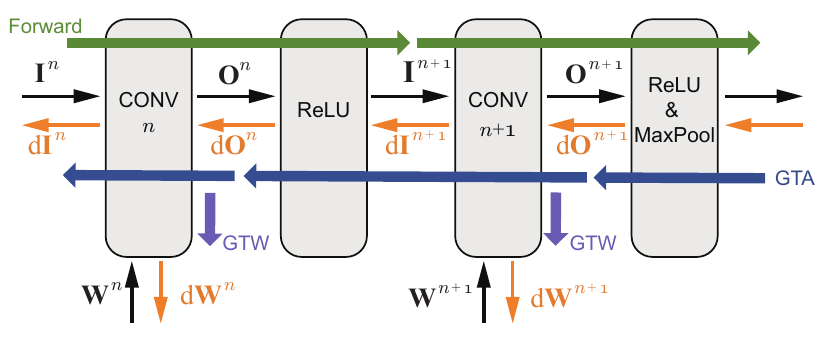}
  \caption{CNN training demonstration.} \label{fig-background-train}
\vspace{-1.2pc}
\end{figure}

A typical CNN training procedure is shown in \cref{fig-background-train}. It contains three stages: Forward, Backward and Weight Update.

\textbf{Forward} stage starts from input layer until final layer, including convolutional (\texttt{CONV}), \texttt{ReLU} and \texttt{MaxPool} layer.
Input activations $\mathbf{I}$ of $n$-th \texttt{CONV} layer are formulated as a 3-D tensor.
The size of $\mathbf{I}$ is determined by the number of input channels $C$, height $H$ and width $W$, and we denote the input activations of $i$-th channel as $\mathbf{I}_i$. Output activations $\mathbf{O}$ are also formulated as a 3-D tensor with $F$ output channels. Weights are formulated as a 4-D tensor with size of $K \times K \times C \times F$, where each convolution kernel $\mathbf{W}_{i, j}$ is a 2-D tensor with the size of $K \times K$. Vector $\mathbf{b}$ is the bias applied on $\mathbf{O}$ after convolution. If 2-D convolution is denoted by ``$\ast$", the \texttt{CONV} layer can be represented as
\[
  \mathbf{O}_i = \sum_{j = 0}^{C} {\mathbf{W}_{i, j}} \ast \mathbf{I}_j + \mathbf{b}_i, \quad
  i = 0, \cdots, F.
\]

\texttt{ReLU} and \texttt{MaxPool} are non-linear operation layers.
\texttt{ReLU} layer applies point-wise function $f(x)=max(0, x)$ on all activations.
\texttt{MaxPool} layer select the maximum value from each window of activations as output.
Hence, the activations usually become sparse after \texttt{ReLU} and \texttt{MaxPool} layers.
And the non-zero patterns generated by \texttt{ReLU} and \texttt{MaxPool} layers are recorded as mask and will be adopted in backward stage.

\textbf{Backward} stage includes two steps:
\begin{itemize}

\item \textbf{Gradient To Activations (GTA)}: it calculates the activation gradients (derivatives to certain layer's activations). According to chain rule, the gradients are calculated from loss function back to input layer. The \texttt{CONV} layer in GTA step is represented as
\[
  {\mathrm{d}\mathbf{I}}_j = \sum_{i = 0}^{F} {\mathrm{d}\mathbf{O}_i \ast {\mathbf{W}^+}_{i, j}}, \quad
  j = 0, \cdots, C,
\]
where $\mathrm{d}\mathbf{I}_j$ indicates the input activation gradients (derivatives of input activations) for the $j$-th channel,
$\mathrm{d}\mathbf{O}_i$ represents the output activation gradients for the $i$-th channel,
$\mathbf{W}_{i, j}$ is the $i$-th filter of $j$-th channel,
$\mathbf{W}^+_{i, j}$ is sequentially reversed of $\mathbf{W}_{i, j}$, or in other words, $\mathbf{W}_{i, j}$ rotated by $180$ degrees.
\texttt{ReLU} and \texttt{MaxPool} layers in GTA step directly adopt the pre-stored mask in forward stage.

\item \textbf{Gradient To Weights (GTW)}: it calculates the weight gradients ${\mathrm{d}\mathbf{W}}$ (derivatives of loss function to layers' weights). These weight gradients are utilized to update weights by Stochastic Gradient Descent (SGD) method. Weight gradients are calculated by
\[
  {\mathrm{d}\mathbf{W}}_{i, j} = {\mathrm{d}\mathbf{O}}_i \ast \mathbf{I}_j, \quad  i = 0, \cdots, F, ~ j = 0, \cdots, C,
\]
where ${\mathrm{d}\mathbf{W}}_{i, j}$ is the weight gradients of $j$-th channel in $i$-th filter.
\end{itemize}

\textbf{Weight Update} stage is to update the weights of each layer with the calculated weight gradients by using SGD method. For modern CNNs, batch training is a popular way to update the weights by sending a batch of inputs (e.g. 32 images) into the network. The weight gradients are computed by averaging the batch of gradients. Finally, the weights are updated according to a pre-set learning rate $\alpha$.
Generally, weight update stage is not a performance bottleneck for CNN training. Thus, \textbf{only the Forward, GTA and GTW procedures are taken into considerations for acceleration}.

\section{Algorithm} \label{section-3}

\subsection{Stochastic Pruning}

\begin{figure}[ht]
  \centering
  \includegraphics[width=0.6\columnwidth]{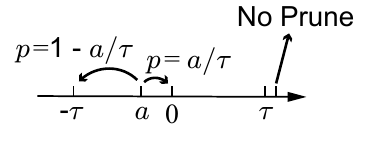}
  \caption{Stochastic gradients pruning algorithm. $\tau$ is the threshold that determines whether a value should be pruned. $p$ is the probability of setting a value to $0$ or $\pm\tau$.} \label{fig:prune}
\end{figure}

From the experiments, we found that there are many gradients data whose absolute value is very small.
Intuitively, a single small gradient value has little impact on the updating of weight, thus it can be pruned (set to zero) directly. However, if many values are pruned, the distribution of gradients is changed remarkably, which causes accuracy loss.
Thus, we adopt a stochastic pruning algorithm proposed by \cite{ye2019accelerating} to solve this problem.

This algorithm treats the gradients that to be pruned (denoted as $g$) as a $n$-dimensional vector, and prunes the vector component whose absolute value is smaller than the threshold ${\tau}$. \Cref{fig:prune} shows the stochastic pruning method. By setting values to zero or $\pm\tau$ stochastically, the expectation of each component remains unchanged, which improves the convergence and accuracy. Detailed analysis can be found in \cite{ye2019accelerating}.


\begin{algorithm}[ht]
  \caption{Overall Pruning Scheme} \label{algo:1}
  \KwIn{$\left[G_1, G_2, ..., G_N\right]$: $N$ batch original activation gradients,
        $N_F$: depth of FIFO, $p$: target sparsity.}
  \KwOut{sparse activation gradients $\left[\hat G_1, \hat G_2, ..., \hat G_N \right]$ }
  $F$ $\triangleq$ FIFO with depth $N_F$.

  \For{$i=1; i \le N ; i \leftarrow i+1$}
  {
 $g \triangleq G_i$ ; $\hat g \triangleq \hat G_i$ \;
 $n =$ length of $g$ ;  $A = 0$ ;

  \For{$j=1;j \le n;j \leftarrow j+1$}
  {

        $A \leftarrow A + \left|g_i\right|$ \;

        \If{$i > N_F$}
        {
            $\hat{\tau}$ = mean$\left(F\right)$ ;

            \eIf{${|g_i| < \hat \tau}$}
            {
                Generate a random number $r \in \left[ {0,1} \right]$ \;
                \eIf{$|g_i| > \hat{\tau} r$}
                {
                    ${\hat g_i} = ({g_i} > 0) $ ? ${\hat{\tau}}$ : $({-\hat{\tau}})$ \;
                }
                {
                ${\hat g_i} = 0 $ \;
                }
            }
            {
                ${\hat g_i} = {g_i} $
            }
        }

  }
  $\tau = {\Phi^{-1}}\left( {\frac{1 - p}{2}} \right) \frac{1}{n}\sqrt {\frac{2}{\pi }} A$ \;
  $F.$push $\left( \tau \right)$ \;
}
\end{algorithm}

\subsection{Threshold Determination and Prediction}

Clearly, it's unfeasible to find the threshold by sorting, due to its huge memory and time consumption. Thus, we propose a hardware friendly threshold selection scheme with less overhead. This selection method contains 2 steps: \textbf{Determination} and \textbf{Prediction}, where the \textbf{Determination} step refers to \cite{ye2019accelerating}.

\begin{figure}[t]
  \centering
  \begin{minipage}[ht]{8pc}
    \centering
    \includegraphics{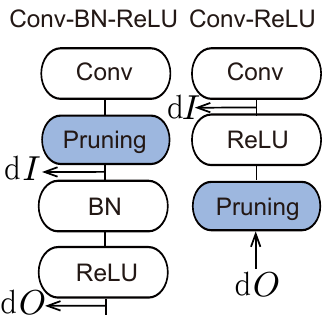}
    \caption{Pruning positions of two typical structures.} \label{fig-struct-type}
  \end{minipage}
  \hfill
  \begin{minipage}[ht]{12pc}
    \centering
    \includegraphics{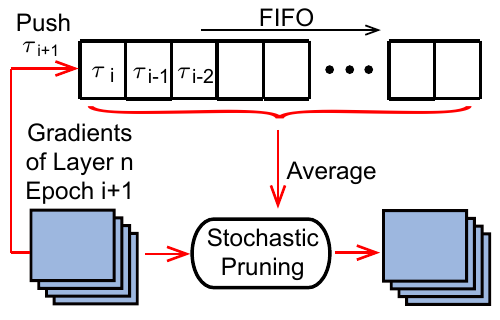}
    \caption{The threshold prediction scheme for each \texttt{CONV} layer using FIFO.} \label{fig-predict}
  \end{minipage}
  \vspace{-1pc}
\end{figure}

\textbf{Threshold Determination.}
Modern CNN models have two typical structures, as shown in \cref{fig-struct-type}.
For the \texttt{CONV-ReLU} structure, where a \texttt{CONV} layer is followed by a \texttt{ReLU} layer, output activation gradients $\mathrm{d}\mathbf{O}$ are sparse, but subject to a irregular distribution.
On the other hand, the input activation gradients $\mathrm{d}\mathbf{I}$, which going to be propagated to the previous layer, is full of non-zero values.
Statistics show that the distribution of $\mathrm{d}\mathbf{I}$ is symmetric with 0 and its density decreases with the increment of absolute value.
For the \texttt{CONV-BN-ReLU} structure, $\mathrm{d}\mathbf{O}$ subjects to the same distribution of $\mathrm{d}\mathbf{I}$.
Simply, we assume these gradients all subject to a normal distribution with mean $0$ and variance ${{\sigma^2}}$ according to \cite{ye2019accelerating}.

In the first case, $\mathrm{d}\mathbf{O}$ can inherit the sparsity of $\mathrm{d}\mathbf{I}$ because ReLU layer won’t reduce sparsity. Thus $\mathrm{d}\mathbf{I}$ can be treated as pruning target $g$ in \texttt{CONV-ReLU} structure.
Besides, in \texttt{CONV-BN-ReLU} structure, $\mathrm{d}\mathbf{O}$ is considered as pruning target $g$.
In this way, the distribution of $g$ in both situations could be unified to normal distribution.

Suppose the length of $g$ is $n$,
the standard deviation $\sigma$ is estimated in a unbiased way \cite{ye2019accelerating} using
\[ \hat \sigma  = \frac{1}{n}\sqrt {\frac{2}{\pi }} \sum\limits_{i = 1}^n {\left| {{g_i}} \right|} \ , \quad {g_i} \in g .\]

Then we can compute the threshold ${\tau}$ using the cumulative distribution function ${\Phi}$ of the standard normal distribution, target pruning rate $p$ and ${\hat \sigma}$:
\[
\tau {\rm{ = }}{\Phi^{-1}}\left( {\frac{1 - p}{2}} \right)\hat \sigma .
\]

\textbf{Threshold Prediction.}
  The stochastic pruning scheme mentioned above needs to access all gradients two times: for computing $A$ and for gradients pruning.
  What is worse, gradients need to be stored in memory temporarily before pruned, which brings overhead of memory access.
  The best way is to prune gradients before they are sent back to memory.
  To accomplish this, we improve the algorithm by predicting the threshold before calculating.
  Denoting the number of batches as $N$, the prediction method keeps a FIFO with a length of $N_F$ for each \texttt{CONV} layer, where $N_F$ is a hyper-parameter that satisfies $N_F << N$.

  \Cref{fig-predict} shows the pruning algorithm with threshold prediction for each \texttt{CONV} layer.
  Activation gradients of each batch will be pruned under the predicted threshold ${\tau'}$ as soon as they are calculated, where ${\tau'}$ is the average of all the thresholds stored in the FIFO.
  At the end of each batch, The determined threshold of this batch is calculated and pushed into the FIFO.
  Gradients will not be pruned before the FIFO is filled up. Details of the whole gradients pruning algorithm is shown in Algorithm~\ref{algo:1}.

  The arithmetic complexity of our algorithm is $\mathcal{O}\left( n \right)$, while complexity of sorting is at least $\mathcal{O}\left( n \log n \right)$. Besides, all the gradients will be accessed just one time in our algorithm, so that almost no extra storage is required, which saves time and energy consumption.

\section{Dataflow} \label{section-4}

\begin{figure*}[ht]
  \centering
  \begin{subfigure}{11.82pc}
    \includegraphics{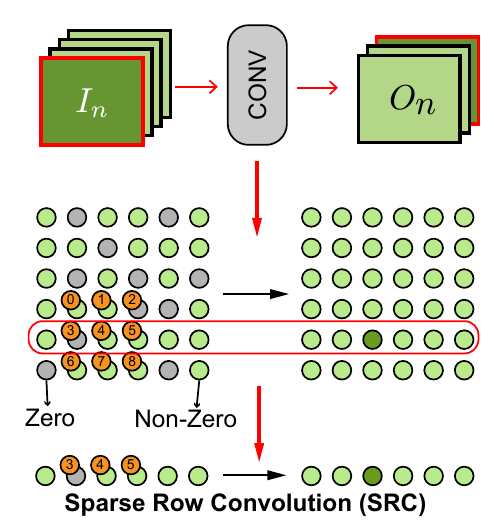}
    \caption{The Forward step.} \label{fig-decomposition-inference}
  \end{subfigure}
  \hfill
  \begin{subfigure}{18.01pc}
    \includegraphics{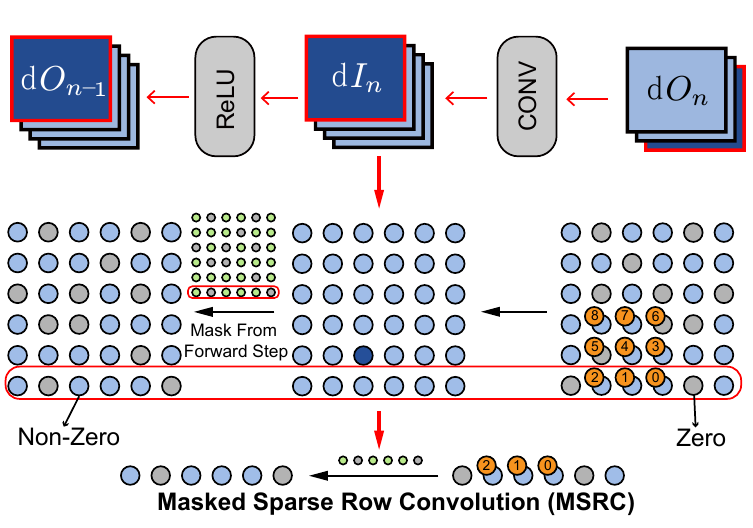}
    \caption{The GTA step.} \label{fig-decomposition-GTA}
  \end{subfigure}
  \hfill
  \begin{subfigure}{11.68pc}
    \includegraphics{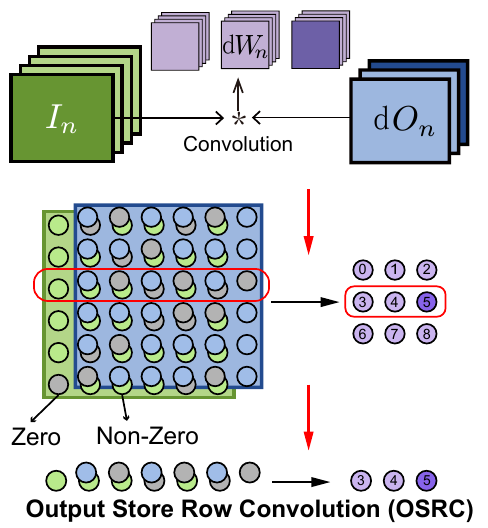}
    \caption{The GTW step.} \label{fig-decomposition-GTW}
  \end{subfigure}
  \caption{Demonstration of our dataflow. Operations in \texttt{CONV} layers are disassembled into channel level and row level. Row level operations are picked as the basic operations of the dataflow.} \label{fig-decomposition}
 \vspace{-1pc}
\end{figure*}

To gain more benefits from the above algorithm, it's essential to design an accelerator that can utilize both activation sparsity and gradient sparsity.
Prior works have shown that the utilized dataflow usually affects the architecture's performance significantly \cite{chen2016eyeriss}.
Thus, we first introduce the dataflow used in the accelerator.

We propose a sparse training dataflow by dividing all computation into 1-Dimensional convolutions. This dataflow supports all kinds of sparsity in training and provides opportunities for exploiting all types of data reuse. This section will introduce the sparsity in training and the dataflow in detail.

\subsection{Data Sparsity in Training} \label{subsection:sparsity}

The sparsity of involved six data types during training have been summarized in \cref{tab-data-sparsity}.
Input activations $\mathbf{I}$ for each \texttt{CONV} layer are usually sparse because the previous \texttt{ReLU} layer sets negative activation values into zeros. Weights $\mathbf{W}$ are also dense for all steps of training.
The output activation gradients $\mathrm{d}\mathbf{O}$ are usually sparse. But for networks with \texttt{BN} layers, $\mathrm{d}\mathbf{O}$ becomes dense after passing through \texttt{BN} layers. However, this issue can be resolved by our gradients pruning algorithm. Thus, we can regard output activation gradients of all \texttt{CONV} layers as sparse data.

There is an additional optimization opportunity in the GTA step. The gradients $\mathrm{d}\mathbf{I}$ are usually sent to a \texttt{ReLU} layer after generated from a \texttt{CONV} layer, leading to certain values be set to zero forcefully. Actually, we could predict the positions of these zeros and skip the corresponding calculations, according to the mask generated in the Forward step.
Besides, both operands (input activations $\mathbf{I}$ and gradients to output activation $\mathrm{d}\mathbf{O}$) in the GTW step are sparse. Theoretically, it could significantly reduce the computation cost in obtaining the weight gradients $\mathrm{d}\mathbf{W}$.

\subsection{Sparse Training Dataflow}

Aiming to utilize all the sparsity in the training process, we divide one 2-D convolution into a series of 1-D convolutions and treat the 1-D convolutions as basic operations for scheduling and running. In this subsection, we will show the way to disassemble three basic steps of CNN training into 1-D convolutions in detail.

\textbf{Forward step.}
As shown in \cref{fig-decomposition-inference}, for one particular 2-D convolution, a small kernel moves through the input activations and perform multiplications and accumulations on each position. More detailed, one output row is the addition of $K$ 1-D convolution results, where $K$ is the kernel size.
For a 1-D convolution in the Forward step, one operand is a certain row of the kernel, which is a short dense vector. Another operand is a row of the input activations, which is a long sparse vector. This basic operation is denoted as Sparse Row Convolution (SRC).

\begin{table}[h]
  \centering
  \caption{Sparsity summary for involved data in training.} \label{tab-data-sparsity}
  \begin{tabular}{|c|c|c|}
    \hline
    Data Type                           & Symbol                   & Sparsity   \\ \hline
    Weights                             & $\mathbf{W}$             & dense      \\ \hline
    Weight Gradients                    & $\mathrm{d}\mathbf{W}$   & dense      \\ \hline
    Input Activations                   & $\mathbf{I}$             & sparse     \\ \hline
    Gradients to Input Activations      & $\mathrm{d}\mathbf{I}$   & dense     \\ \hline
    Output Activations                  & $\mathbf{O}$             & dense     \\ \hline
    Gradients to Output Activations     & $\mathrm{d}\mathbf{O}$   & sparse     \\ \hline
  \end{tabular}
\end{table}

\textbf{GTA step.}
\Cref{fig-decomposition-GTA} shows the decomposition of the GTA step.
Similar to the Forward step, the GTA step can also be regarded as the summation of multiple 1-D convolutions.
However, as mentioned in \cref{subsection:sparsity}, certain gradients are set to zero forcefully after sent to \texttt{ReLU} layers. Thus, the calculation of these masked values are entirely unnecessary and can be skipped safely to save computation cost. Considering this optimization, we define a different basic operation named as Masked Sparse Row Convolution (MSRC).

\textbf{GTW step.}
The weight gradients computation in the GTW step is shown in \cref{fig-decomposition-GTW}.
The GTW step is significantly different from the Forward and the GTA step in two ways.
The first is that operands of extracted 1-D convolutions are two sparse long vectors.
The second lies in the range of results:
traditional convolutions need to slide one vector from the start of the other vector to its end, and calculating multiplications and accumulations in each sliding position. However, in the GTW step, only the results in several positions are needed, and it is not necessary to fully calculate all the convolution results.
The resulting vector is usually short (as the kernel size $K$) and we can store it in a small scratchpad during the whole convolution. Thus, this 1-D convolution is named Output Store Row Convolution (OSRC) and is considered as the basic operations of the GTW step.

Gradients to bias in \texttt{CONV} layers are also required in the GTW step. The calculation of bias gradients for each channel is just the summation of the output activation gradients from the corresponding channel. It is simple to calculate them by accumulating gradients during the GTA step.

\section{Architecture} \label{section-5}

\begin{figure}[t]
\centering
\hfill
\begin{subfigure}{11pc}
  \centering
  \includegraphics{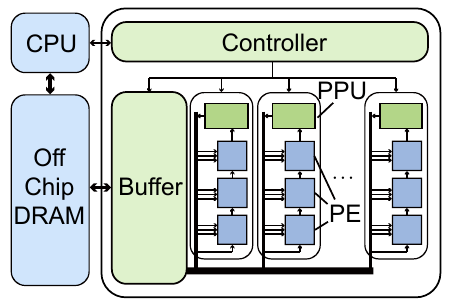}
  \caption{Overview architecture} \label{fig-architecture-overview}
\end{subfigure}
\hfill
\begin{subfigure}{6.8pc}
  \centering
  \includegraphics{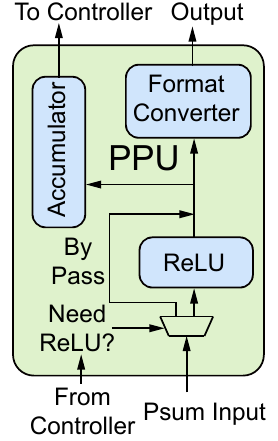}
  \caption{PPU module} \label{fig-architecture-ppu}
\end{subfigure}
\hfill
\par
\begin{subfigure}{14.2pc}
  \centering
  \includegraphics{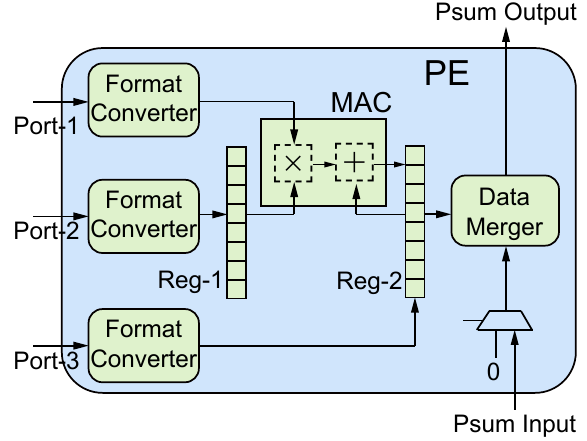}
  \caption{PE module} \label{fig-architecture-pe}
\end{subfigure}
\caption{The architecture design of \textit{SparseTrain}.}
\vspace{-1pc}
\end{figure}

\begin{table*}[ht]
  \centering
  \begin{threeparttable}
    \caption{Evaluation results for the gradients pruning algorithm, where \texttt{acc}\% means the training accuracy and $\rho_{nnz}$ means the density of non-zeros.} \label{tab:spar-acc}
    \begin{tabular}{cc|cc|cc|cc|cc|cc}
      \specialrule{0.8pt}{0pt}{0pt}
      \multirow{2}{*}{Model}&
      \multirow{2}{*}{Dataset}&
      \multicolumn{2}{c|}{Baseline}&
      \multicolumn{2}{c|}{$p=70\%$}&\multicolumn{2}{c|}{$p=80\%$}&\multicolumn{2}{c|}{$p=90\%$}&\multicolumn{2}{c}{$p=99\%$} \\
      \cline{3-12}
      & & \texttt{acc}\% & $\rho_{nnz}$ & \texttt{acc}\% & $\rho_{nnz}$ & \texttt{acc}\% & $\rho_{nnz}$ & \texttt{acc}\% & $\rho_{nnz}$ & \texttt{acc}\% & $\rho_{nnz}$   \\
      \hline
      AlexNet & CIFAR-10&90.50&0.09&90.34&0.01&\textbf{90.55}&0.01&90.31&0.01&89.66&0.01 \\
      ResNet-18 & CIFAR-10&95.04&1&\textbf{95.23}&0.36&95.04&0.35&94.91&0.34&94.86&0.31 \\
      ResNet-34 & CIFAR-10&94.90&1&95.13&0.34&95.09&0.32&\textbf{95.16}&0.31&95.02&0.28 \\
      ResNet-152 & CIFAR-10&\textbf{95.70}&1&95.13&0.18&95.58&0.18&95.45&0.16&93.84&0.08 \\
      AlexNet & CIFAR-100&67.61&0.10&67.49&0.03&\textbf{68.13}&0.03&67.99&0.03&67.93&0.02 \\
      ResNet-18 & CIFAR-100&76.47&1&76.89&0.40&\textbf{77.16}&0.39&76.44&0.37&76.66&0.34 \\
      ResNet-34 & CIFAR-100&77.51&1&77.72&0.36&\textbf{78.04}&0.35&77.84&0.33&77.40&0.31 \\
      ResNet-152 & CIFAR-100&79.25&1&\textbf{80.51}&0.22&79.42&0.19&79.76&0.18&76.40&0.10 \\
      AlexNet & ImageNet&56.38&0.07&\textbf{57.10}&0.05&56.84&0.04&55.38&0.04&39.58&0.02 \\
      ResNet-18 & ImageNet&68.73&1&\textbf{69.02}&0.41&68.85&0.40&68.66&0.38&68.74&0.36 \\
      ResNet-34 & ImageNet&\textbf{72.93}&1&72.92&0.39&72.86&0.38&72.74&0.37&72.42&0.34 \\
      \specialrule{0.8pt}{0pt}{0pt}
    \end{tabular}
  \end{threeparttable}
\vspace{-0.6pc}
\end{table*}

Aiming to evaluate the dataflow sparsity in \textit{SparseTrain}, we design an architecture of which overview is shown in \cref{fig-architecture-overview}.
The architecture consists of Buffer, Controller, and PE groups. Each PE group contains $3$ PEs and a Post Processing Unit (PPU). PE is designed to calculate 1-D convolution and PPU is adopted for point-wise operations.

\textbf{PE Architecture}:
\label{subsection:pe}
As demonstrated in \cref{section-4}, there are three kinds of basic operations in our dataflow: SRC, MSRC and OSRC. PE module is designed to support all of these operations and the architecture of PE module is shown in \cref{fig-architecture-pe}.

Our PE is designed to perform a complete 1-D convolution, instead of just one multiplication.
Each time an input value is loaded from Port-1, PE multiplies it by $K$ values in Reg-1, providing $K$ product results stored in Reg-2.

When performing SRC operations, A PE first loads weight vector from Port-2 and saves them in Reg-1. Then activation values are loaded from Port-1 and multiplied by the weights in Reg-1. Results are accumulated to Reg-2. This process repeats until the activation vector gets to the end.

For MSRC operations, the offset vector of input activations ($\mathbf{I}$) is loaded from Port-3 and saved to Reg-2. Values in the offset vector is used to indicate the results that should be calculated. In other words, values that are not encoded in the offsets of $\mathbf{I}$ can be predicted as zero and skipped safely. During calculation, PE will look ahead for the next non-skip operand from Port-1, so that it can start computation as soon as possible without extra cycles waiting for input data.

For OSRC operations, PE stores weight gradients ($\mathrm{d}\mathbf{W}$) in Reg-2 during computing. $\mathbf{I}$ and $\mathrm{d}\mathbf{O}$ vector are loaded from Port-1 and Port-2, respectively. During computing, $K$ values of $\mathrm{d}\mathbf{O}$ will be cached in Reg-1 and multiplied with each value in $\mathbf{I}$. The results will be accumulated to Reg-2.

\textbf{PPU Architecture}:
In general, PPU is responsible for all point-wise operations. \Cref{fig-architecture-ppu} shows the internal architecture of PPU.
In PPU, resulting vector will be converted into a compressed format and sent back to buffer. PPU will also perform \texttt{ReLU} operation before format conversion if needed. During GTA step, all of the gradients through PPU modules and their absolute value will be accumulated into two registers respectively. Accumulation results are used for generating bias gradients and determining pruning threshold.

\section{Evaluation} \label{section-6}

In this section, several experiments are conducted to demonstrate that the proposed approach could reduce the training complexity significantly with a negligible model accuracy loss.
In the following experiments, AlexNet \cite{krizhevsky2012imagenet} and ResNet \cite{he2016deep} series are evaluated on three datasets including CIFAR-10, CIFAR-100 and ImageNet. All the models are trained for $300$ epochs on CIFAR-\{10, 100\} and $180$ epochs on ImageNet, due to our limited computing resources. PyTorch framework is adopted for the algorithm verification.

To verify the performance of our dataflow and architecture, a custom cycle-accurate C{++} simulator is implemented based on SystemC model for the proposed architecture, and a simple compiler is designed with Python to convert CNN models in PyTorch into our internal instructions for driving our simulator. We also have the RTL implementation for PE, PPU and controller, which is synthesized by Synopsys Design Compiler with Global Foundries $14 nm$ FinFET technology for area estimation and then simulated by Synopsys PrimeTime for power estimation. The area and energy consumption of buffer (SRAM) is estimated by PCACTI \cite{shafaei2014fincacti}.
To demonstrate the advantages of our proposed sparse training dataflow, we compare it with the architecture from Eyeriss \cite{chen2016eyeriss}. Since Eyeriss is designed for CNN inference rather than training, we modify the architecture of Eyeriss to support the dense training process.
We adopt $168$ PEs in both the proposed architecture and the baseline architecture. For convenience, $386$KB SRAM is utilized as the global buffer for intermediate data, which is sufficient for storing data used in each iteration. A larger buffer is beneficial to improving data-reuse and energy efficiency, but it is beyond the considerations of this work.


\subsection{Sparsity and Accuracy}

From \cref{tab:spar-acc}, there is no accuracy lost for most situations and even $1\%$ accuracy improvement for ResNet-50 on CIFAR-100.
The only significant accuracy lost exists in AlexNet on ImageNet, when using a very aggressive pruning policy like $p = 99.5\%$.
That proves the accuracy loss caused by our layer-wise gradients pruning algorithm is almost negligible.

The gradient density illustrated in \cref{tab:spar-acc} shows that our method could reduce the gradient density by $3\times \sim 10\times$. In addition, the deeper networks could obtain a relatively lower gradient density with our sparsification, which means that our method works better for larger networks.

\begin{figure}[t]
  \centering
  \includegraphics{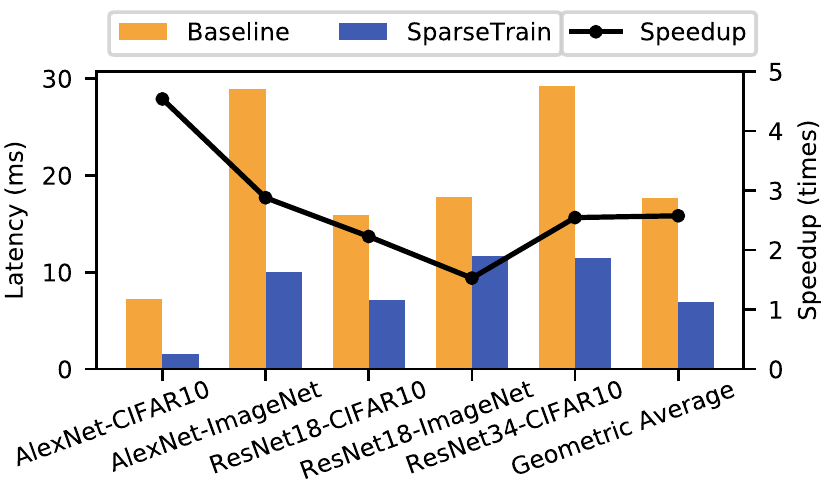}
  \caption{Average training latency per sample for different models and datasets. The speedup is the training latency of \textit{SparseTrain} compared with the baseline.} \label{fig-evaluate-time}
\vspace{-1pc}
\end{figure}
\begin{figure}[t]
  \centering
  \includegraphics{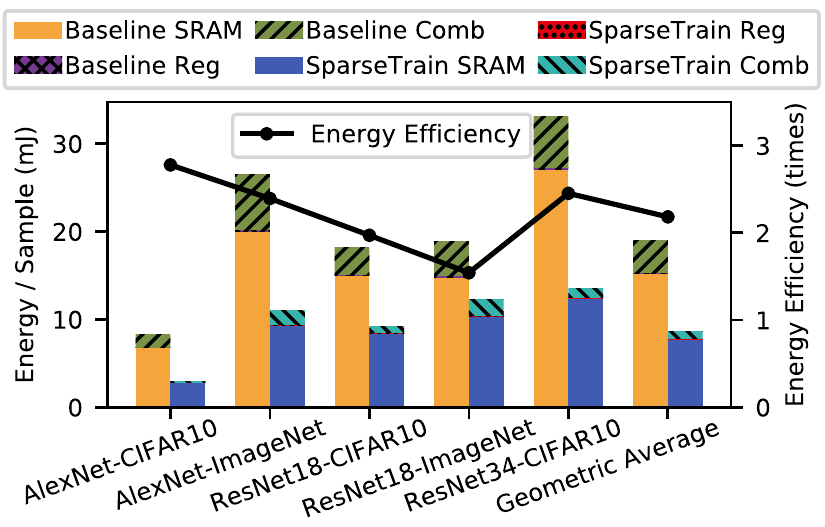}
  \caption{Average energy consumption per sample for different models and datasets. ``Reg'' represents register, and ``Comb'' represents combinational logic. The energy efficiency is the improvement of \textit{SparseTrain} compared with the baseline.} \label{fig-evaluate-energy}
\vspace{-1pc}
\end{figure}

\subsection{Convergency}

The training loss of AlexNet/ResNet-18 on CIFAR-10 and ImageNet is demonstrated in \cite{ye2019accelerating}.
In general, ResNet-18 is very robust for gradients pruning.
For AlexNet, the gradients pruning could still be robust on CIFAR-10 but will degrade convergency speed a little on ImageNet with aggressive pruning policy. These results prove that with reasonable hyper-parameter $p$, our pruning algorithm basically has the same convergency property compared with the original training scheme.

\subsection{Latency and Energy}

\Cref{fig-evaluate-time} shows the training latency reduction brought by sparsity exploration.
The proposed \textit{SparseTrain} scheme achieves $4.5 \times$ speedup at most for AlexNet on CIFAR-10. On average, it achieves about $2.7 \times$ speedup compared with the baseline.

\Cref{fig-evaluate-energy} shows the average energy consumption per data sample.
Overall, \textit{SparseTrain} has $1.5 \times$ to $2.8 \times$ (on average $2.2 \times$) energy efficiency improvement than baseline.
For the baseline, $62\% \sim 71\%$ of the energy consumption comes from SRAM accesses. \textit{SparseTrain} reduces the global buffer accesses by utilizing sparse dataflow and reduces the energy cost by $30\% \sim 59\%$.
The energy consumption of combinational logic in \textit{SparseTrain} could be reduced by $53\% \sim 88\%$, which is more significant than SRAM and register accesses.
This also contributes much to the total energy saved.

\section{Conclusion} \label{section-7}

As the model size and datasets scale larger, CNN training becomes more and more expensive. In response, we propose a novel CNN training scheme called \textit{SparseTrain} for acceleration by fully exploiting sparsity in the training process.
By performing stochastic gradients pruning on each layer, \textit{SparseTrain} achieves high sparsity with negligible accuracy loss.
Additionally, with a 1-D CONV based sparse training dataflow and a sparse-aware architecture, \textit{SparseTrain} can improve CNN training speed and efficiency significantly by exploiting different types of sparsity. With the threshold prediction method, gradients pruning can be performed on our architecture with almost no overhead.
Experiments show that the gradients pruning algorithm could achieve $3\times$ to $10\times$ sparsity improved with negligible accuracy loss. Compared with the baseline, \textit{SparseTrain} achieves about $2.7 \times$ speedup and $2.2 \times$ energy efficiency improvement on average.

\small

\begingroup

\bibliographystyle{IEEEtran}


\endgroup

\end{document}